\documentclass{article}
\usepackage{graphicx}
\usepackage[margin=1in]{geometry}
\usepackage[T1]{fontenc}
\usepackage[utf8]{inputenc}
\usepackage{lmodern}
\usepackage{microtype}
\usepackage{booktabs}
\usepackage{array}
\usepackage{calc}
\usepackage{caption}
\usepackage{amsmath}
\usepackage{longtable}
\usepackage{CJKutf8}
\usepackage{hyperref}
\newcommand\tightlist{\setlength{\itemsep}{0pt}\setlength{\parskip}{0pt}}
\hypersetup{
  pdftitle={The interpreter as developer: An autoethnographic design study of building a computer-assisted interpreting tool through human--LLM co-creation},
  pdfauthor={Zhao Zeming},
  hidelinks,
  pdfcreator={LaTeX via pandoc}}

\title{The interpreter as developer: An autoethnographic design study of
building a computer-assisted interpreting tool through human--LLM
co-creation}
\author{Zhao Zeming}
\date{}

\begin{document}
\maketitle
\begin{CJK}{UTF8}{gbsn}

\textbf{Abstract}

Computer-assisted interpreting (CAI) tools are built for interpreters,
mainly by third parties, and the interpreters' own side of development
has received little first-person documentation. This article reports an
autoethnographic design study in which a professional conference
interpreter without formal programming training built a local-first CAI
system through expert-led human--LLM co-creation. Analysis of the
development dialogues shows how tacit professional knowledge became
design only after generic defaults had failed in use, a reversal of the
usual direction of technological mediation, from adopting tools to
authoring them. Three mechanisms of in-booth appropriation emerged under
a human-led division of labor, and a four-layer reconstruction
illustrates the human contribution at work. First-month telemetry, over
six thousand utterances in nine sessions, shows a fully local system
sustaining translation-stage latencies near one second, with
sustained-load degradation as a working boundary. The case bears on
professional agency, knowledge production, and the training of
interpreters under generative AI.


\section{1. Introduction}\label{introduction}

Computer-assisted interpreting (CAI) has grown into a small research
field (Fantinuoli 2018, 2023; Chan 2025), but the tools examined there,
systems such as InterpretBank (Fantinuoli 2016) and the VIP platform
(Corpas Pastor 2020), are built for interpreters, most of them by third
parties or by researcher-developers working from the research side of
the field. A first-person, practice-level record of an interpreter
building a tool for their own booth has been lacking. The lack is not
only documentary. When generative systems increasingly mediate how
language work is performed, who holds design authority over a
profession's tools becomes a question about agency itself: a field
served only by tools built elsewhere can state its needs as a consumer,
while a practitioner who builds retains the power to define what the
technology is for. This article examines that shift through the complete
record of one build. In the conferences in which the author works,
practitioners' working reliance falls mainly on an informal,
near-universal assemblage of consumer applications: transcription apps
such as iFlytek, the captioning built into Tencent Meeting and Zoom, and
general-purpose AI assistants such as Yuanbao and Doubao. The streamed
text is sentence-shaped, self-revising, and designed for audience
reading, so intra-sentence content is unstable and settling a phrase
takes longer than the lag allows; the services are online, bringing
network fragility and, more seriously, a standing conflict with the
confidentiality terms of client work. The lag and attention constraints
have long been studied in the effort-model literature (Gile 2009; Seeber
2011); that current consumer services do not meet them is the author's
observation from practice in this market. What the booth needs is terms,
names, and numbers that arrive within the interpreter's natural lag
behind the speaker and hold still once there, on hardware that travels,
with nothing leaving the room.

This article is written from that underdocumented side. The interpreter
whose practice it reports is a freelance professional with over one
hundred conference days per year for more than a decade, holding a
doctorate in the history of science and technology and a master's degree
in translation and interpreting. The interpreter is the author. In the
summer of 2026, after a year of smaller experiments, and once
conversational code generation had made building itself conversational,
a working tool took shape over two weeks of August, with a minimum
viable version in productive use within two hours of the project's first
morning, and the tool has accompanied real assignments since that day.
The study asks three questions of the complete record that build
produced.

\begin{itemize}
\tightlist
\item
  \textbf{RQ1.} How does a professional simultaneous interpreter without
  formal programming training appropriate co-creative development based
  on large language models (LLMs) to design and implement a booth-ready
  tool?
\item
  \textbf{RQ2.} What interpreting-specific tacit knowledge surfaces in
  this process, and through what mechanisms is it codified into concrete
  design decisions?
\item
  \textbf{RQ3.} How does using the self-built tool in a demanding real
  event manifest in the interpreter's real-time strategies and
  experienced cognitive load?
\end{itemize}

The study makes three contributions. It offers, to the author's
knowledge, the first such record in interpreting studies: a complete
first-person account of an interpreter building a working AI tool for
their own booth, with the founding dialogues archived and available for
audit on reasonable request. It maps the professional knowledge that
entered the design, documenting how, in the selected episodes, the
practitioner originated, revised, and accepted each load-bearing
requirement. And it introduces a four-layer documentation framework for
observing human-AI collaboration in real work, illustrated here with one
reconstructed moment from the interpreter's booth.

\section{2. What the booth demands}\label{what-the-booth-demands}

Simultaneous interpreting runs on a small, deliberate delay: the
interpreter renders the speaker's words one to three seconds behind,
with substantial individual variation, a lag known as the ear-voice span
and studied as a measure of cognitive load (Collard and Defrancq 2019;
Janikowski and Chmiel 2025). The first consequence for any display is
timing: support that arrives after the interpreter has committed to a
rendering cannot shape it, so late completeness is worthless, and the
most valuable content, terms, names, and numbers, must arrive within or
near the lag and hold still once there. This is precisely the failure of
the consumer captioning described in the introduction. Arrival inside
the lag is necessary, and so is standing still within it.

The second consequence concerns attention: interpreting consumes most of
the mind's capacity for managing effort, in the terms of the effort
models (Gile 2009; Seeber 2011), and a visual display competes for those
same resources, so anything unstable on the screen imposes exactly the
load the display was meant to relieve. Eye-tracking evidence from
consecutive interpreting shows captioning displays redirecting gaze
(Chen and Kruger 2023, 2024), and experiments report gains from
automatic speech recognition (ASR) number support (Defrancq and
Fantinuoli 2021; Pisani and Fantinuoli 2021) and term support (Prandi
2023). Whether a display that holds still and carries only what the
interpreter needs can serve inside the lag remains, on this evidence, a
design hypothesis the tool was built to meet.

The third consequence is that tools do not merely assist an unchanged
activity; they redistribute where the effort of the activity resides, a
view developed for interpreting by Mellinger (2023) and argued from the
systems side by Fantinuoli (2026). If tools redistribute effort, then
who designs the tool decides where the effort goes, and design decisions
made without knowledge of booth practice may create additional demands
on interpreters. The converse is the case at issue here. When the
practitioner becomes the designer, the redistribution of effort becomes
an exercise of professional agency instead of something imposed on the
profession. The design decisions examined below are, at the same time,
decisions about where cognitive effort resides in mediated interpreting,
taken by the person who bears it.

A fourth constraint arose from the author's practice and ethics rather
than any universal standard: the system runs offline, since booths are
often connectivity-poor and client confidentiality can forbid sending
the event's audio to any external service; in the assignments reported
here the sessions ran entirely locally, the guarantee holding by
configuration rather than policy.

These four demands defined the build standard the tool had to meet, and
the interpreting knowledge at stake is largely tacit, in Polanyi's
(1966) sense: practitioners can act on it before they can say it, and it
becomes usable for design only through the effort of putting it into
words (Nonaka and Takeuchi 1995). The case can also be read through
lead-user theory (von Hippel 1986, 2005): the interpreter faced this
need ahead of the market and stood to benefit directly from a solution,
the two defining attributes of a lead user, and, unserved, built for
themselves, an interpretive lens rather than a tested classification.

\section{3. Method}\label{method}

\subsection{3.1 An autobiographical design
study}\label{an-autobiographical-design-study}

The study is an autobiographical design study with an autoethnographic
analysis. From autobiographical design (Neustaedter and Sengers 2012) it
takes its object, a system designed for the designer's own sustained
use, here working infrastructure relied on in paid assignments. From
analytic autoethnography (Anderson 2006; Chang 2008) it takes its method
of analysis, the systematic qualitative analysis of the author's own
practice records, with the obligation to move beyond narrative toward
concepts that travel. Combining this stance with instrumented
measurement requires care (Hokkanen 2017); the study confines
instrumentation to corroboration. Three features are plain. The primary
instrument of observation is the author, and the study claims analytic
rather than statistical generalization. One person held all three roles
of developer, user, and researcher, so the usual distance against
self-serving description is absent, and the countermeasures are
triangulation against the timestamped logs, retention of negative
episodes, and explicitly bounded claims. And the case is bounded to
bidirectional Chinese--English conference interpreting in the Chinese
market.

\subsection{3.2 The triadic co-creation
workflow}\label{the-triadic-co-creation-workflow}

Development proceeded through a triadic co-creation workflow with three
roles (Figure 1). The domain expert, the author, originates
requirements, supplies professional knowledge, judges every intermediate
result against practice, and performs final acceptance. The LLM
architect (Gemini 3.1 Pro, later GLM-5.3) receives problems stated in
professional language and returns structured build instructions in
Markdown, scoped tasks, file-level changes, and acceptance criteria, a
scaffold converting complaint into engineering specification. The coding
agent (Cursor, later ZCode) implements, refactors, and debugs. The
operative loop ran from a practice-originated need through architect
dialogue and structured instruction to execution, realistic testing, and
deviation feeding back; every instruction records professional knowledge
put into words, so the dialogue logs document knowledge that would
otherwise have stayed tacit.

The triadic configuration describes the founding period; the workflow
did not stand still. When the stack migrated at the end of August, the
architect and executor functions folded into one conversational agent,
and development later organized into a main session with specialized
sub-sessions. Across every configuration the division of labor held:
requirements originated in professional practice, acceptance rested with
the practitioner, and decomposition and execution remained the machine's
contribution.

The label such practice soon acquired, vibe coding (Karpathy 2025), was
coined for conversational programming that accepts output with little
inspection; empirical studies find a spectrum, from such delegation to
selective human verification and testing (Ge et al.~2025; Sarkar and
Drosos 2025; Fawzy et al.~2026). This workflow belongs to the family in
medium, differing in its acceptance regime, governed by professional
practice throughout: expert-led conversational development.

\begin{figure}[htbp]
\centering
\includegraphics[width=0.85\textwidth,height=0.85\textheight,keepaspectratio]{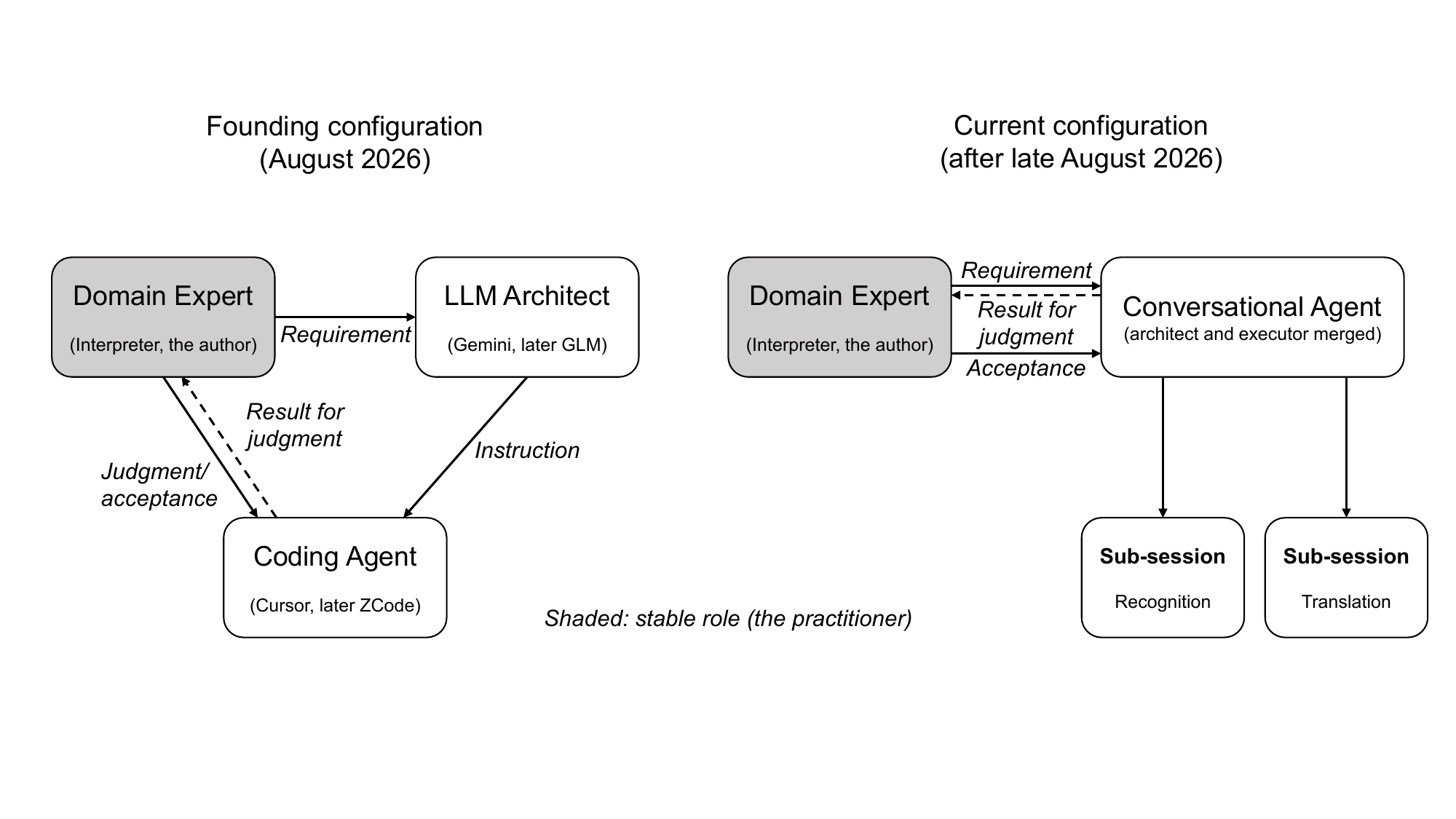}
\caption{The co-creation workflow in its founding and current
configurations. Shaded boxes mark the stable role, the practitioner who
originates requirements and holds acceptance; unshaded boxes mark the
machine side, begun as two products and later folded into one
conversational agent with specialized sub-sessions.}
\end{figure}

\subsection{3.3 Data and analysis}\label{data-and-analysis}

The study assembled five classes of data, development dialogues (a
Gemini thread of 101 exchanges over the founding sprint, plus an
executor-side session of 122 timestamped instructions, 12--27 August
2026), design memos and a reflective journal, a usage record (the tool
entered real booth use on the day the project began), session telemetry
(paired logs per module with timestamp, model, text, and latency, plus a
synchronized recording), and a replay corpus (the focal event's public
broadcast, replayed through the same local chain). Provenance is
documented: the founding records survive in three platform exports
consolidated into a dated, version-controlled retrospective archive, the
working record of the analysis itself. Completeness is claimed against
that window and those platforms.

The data window is stated explicitly: the founding dialogue corpus
closed on 27 August, marking the end of the documented founding build
cycle, and the reported telemetry ends with the September sessions.
Development and daily professional use continued beyond it, the tool
growing from the v0.2 rewrite of late August to v0.6.2 at the time of
writing, publicly archived in the repository, while this article's
analysis remains anchored to the window it claims. Table 1 fixes the
chronology of build events, field deployments, and evidence.

{\def\LTcaptype{none} 
\begin{table}[htbp]
\centering
\caption{Timeline of build events, field deployments, and the analysis window (2026).}
\label{tab:timeline}
\small
\begin{tabular}{p{0.10\linewidth}p{0.42\linewidth}p{0.20\linewidth}p{0.20\linewidth}}
\toprule
Date & Event & Tool state & Primary evidence \\
\midrule
12 Aug & Project begins; minimum viable version in two hours; first live deployment; two-day architect sprint opens (101 exchanges) & v0.1, manual direction & Dialogues; broadcast \\
18 Aug & Focal event of Section 5; fixed 800 ms threshold too strict & v0.1 founding set & Journal; broadcast \\
27--28 Aug & Founding corpus closes; repository initialized; instrumented assignment & v0.1 stable & Exports; telemetry (971) \\
31 Aug--1 Sep & v0.2 sprint; threshold a live control; focal broadcast replayed on the day's chain settings & v0.2 & Commits; replay \\
September & Four conference days, eight sessions & v0.3--v0.4 & Telemetry (5,274) \\
Late Oct & Punctuation-primary segmentation in routine use & v0.5+ & Repository \\
\bottomrule
\end{tabular}
\end{table}
}

The analysis ran concurrently with the design work, and reflexive
thematic analysis (Braun and Clarke 2006) served as the nearest
methodological label rather than a procedure followed phase by phase.
The founding dialogues were exported and read closely in one
consolidated pass at August's end. A working list of major events, kept
alongside the design memos, was consolidated into the mapping of Table
2, and the episodes of Section 4 were then written as narrations of
selected rows of that table. Episodes were segmented by the trigger that
opened each one and selected by two criteria: the failure had to
originate in a professional constraint, not general engineering
friction, and the resolution had to be traceable in the logs.
Implementation-friction episodes such as packaging failed the first
criterion. These criteria governed the four narrated episodes; the
remaining rows of Table 2 entered through code audit, and the recurrence
of professional constraints in the narrated record is partly an artifact
of selection. In practice a failure was recalled first from memory and
then located in the logs, and quotations follow a hierarchy: verbatim
where marked as quoted, working translations where marked as translated,
paraphrase elsewhere. One chain illustrates the procedure. Turn 19 of
the architect thread (12 August) records the requirement in the original
Chinese; a working translation of an excerpt runs, `I do not need each
sentence split at fixed time intervals\ldots{} the longest-sentence time
(a settable 5--30 s) is a ceiling on a single pass', read as a
unit-of-translation judgment: completeness, not clock time, should
release a segment. The decision became Table 2 row 5. The field supplied
the counter-evidence: on 18 August segments rode the cap; the 31 August
sprint made the threshold a live control; by late October punctuation
finalized most segments in routine use. AI sessions assisted with
retrieving and tallying log material under the author's direction, and
the identification of episodes and all interpretive judgment rest with
the author.

\section{4. From expertise to design}\label{from-expertise-to-design}

\begin{figure}[htbp]
\centering
\includegraphics[width=0.85\textwidth,height=0.85\textheight,keepaspectratio]{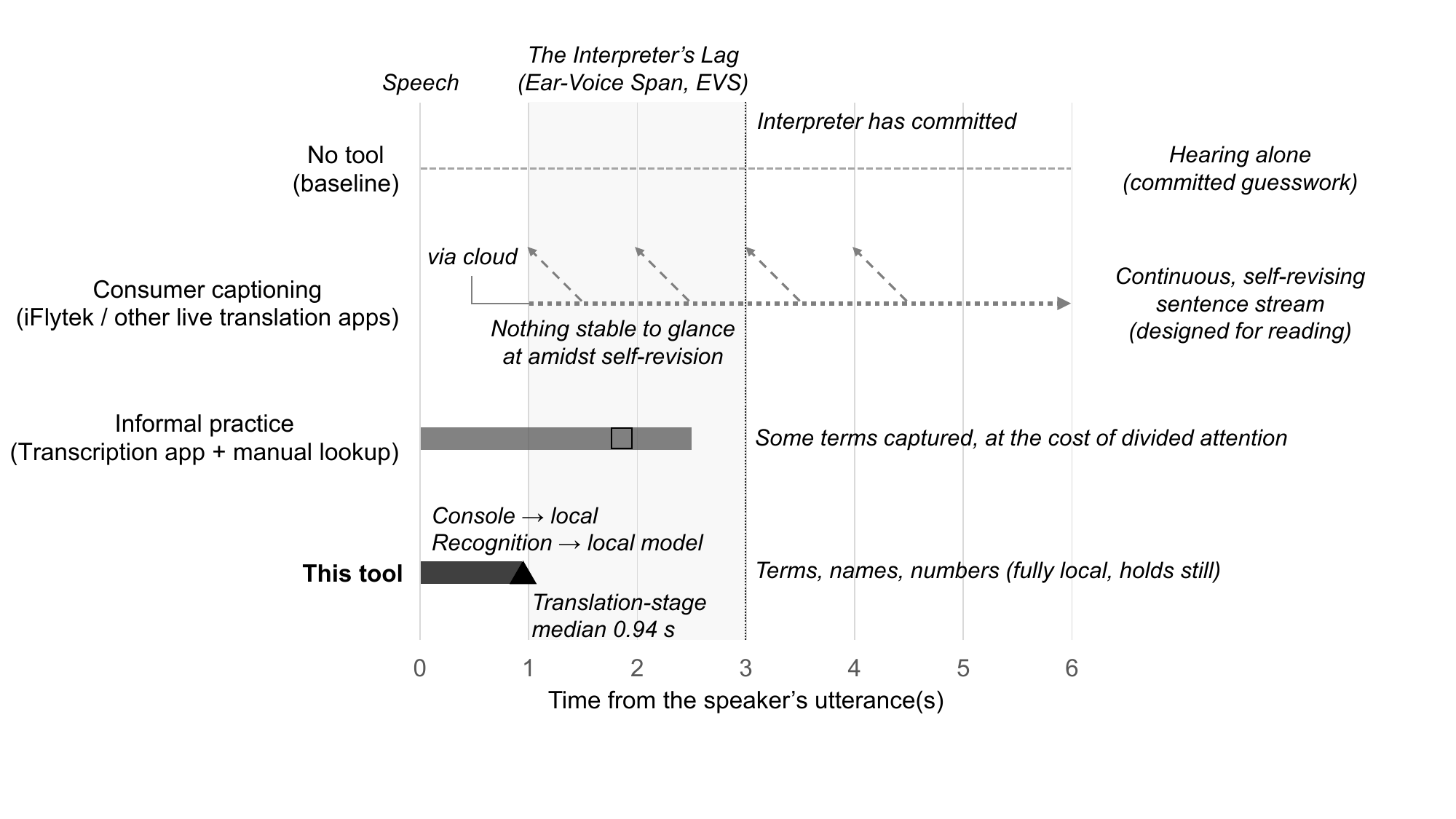}
\caption{Four routes from the same speech. Consumer systems
stream self-revising, audience-designed sentence translation, offering
nothing stable to glance at within the lag. Informal practice combines
transcription apps with manual lookup at the cost of divided attention.
The tool built here runs locally, delivers terms, names, and numbers at
a translation-stage median of 0.94 seconds (28 August), and holds still
once delivered.}
\end{figure}

The outcome is a working system for Chinese--English booth practice on
an ordinary laptop, taking audio from the console's standard headphone
output with consumer signal processing switched off at capture. Speech
recognition and translation run on separate local models, both residing
within the laptop's hardware limits, with an optional cloud route,
unused in the reported sessions, that yields to local processing
whenever the connection fails. Two windows coexist, a full workspace and
a thin subtitle bar that floats above other documents and never sleeps
(the current version adds a third, dockable subtitle window, after the
analysis window).

\begin{figure}[htbp]
\centering
\includegraphics[width=0.85\textwidth,height=0.85\textheight,keepaspectratio]{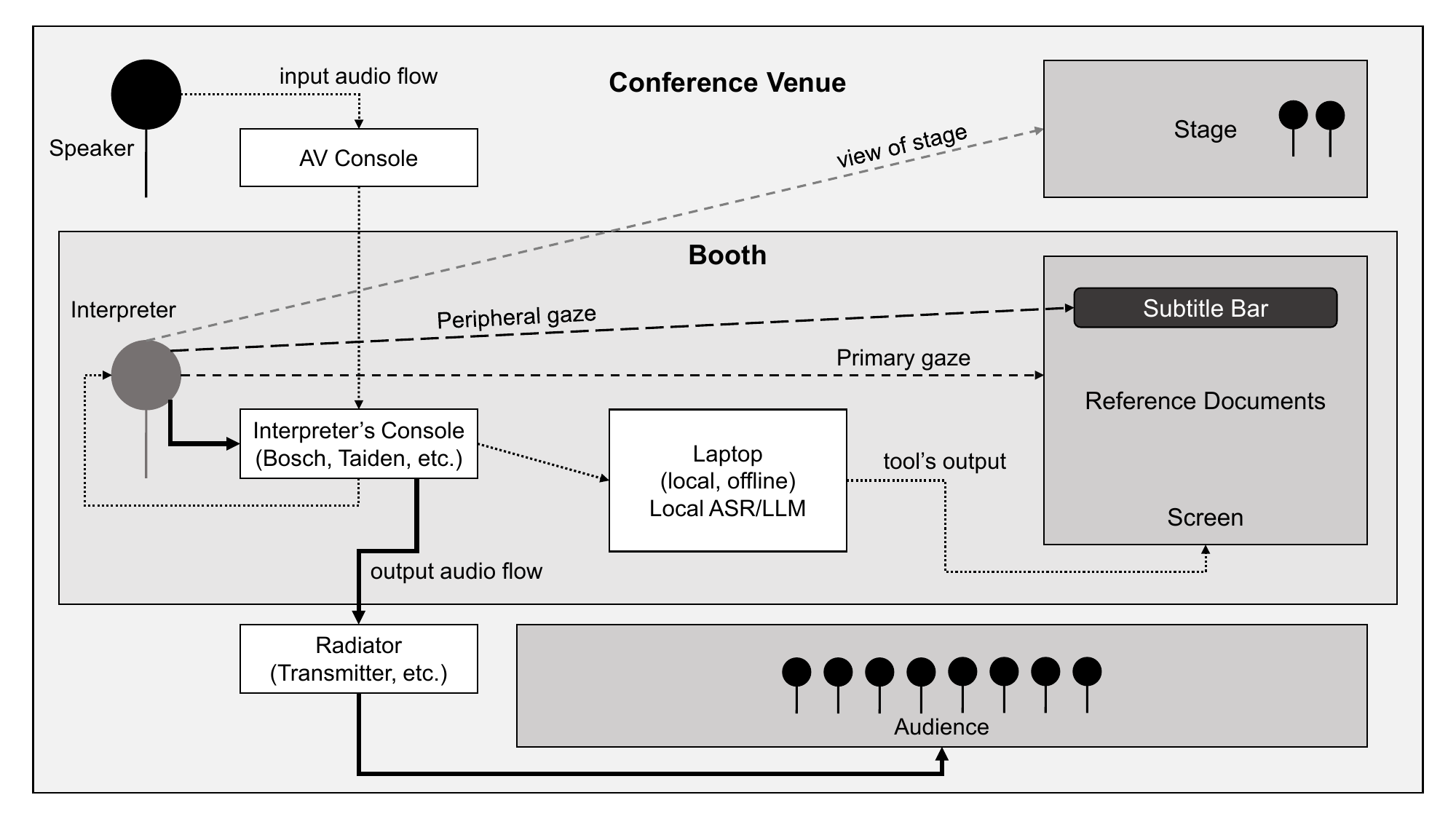}
\caption{The tool as deployed in the booth. Audio is taken from
the console's analogue output through an isolated split, leaving
monitoring untouched; recognition and translation run locally; the
subtitle bar floats above reference documents, read peripherally.}
\end{figure}

The analytic heart is not the system but the knowledge articulated to
obtain it. Table 2 maps each load-bearing design decision to the
professional knowledge that motivated it, and four mappings are narrated
as episodes.

{\def\LTcaptype{none} 
\begin{table}[htbp]
\centering
\caption{Professional knowledge mapped to design decisions in the founding build (12--27 August 2026).}
\label{tab:mapping}
\small
\begin{tabular}{cp{0.22\linewidth}p{0.20\linewidth}p{0.32\linewidth}p{0.14\linewidth}}
\toprule
\# & Knowledge or constraint (practice) & Design requirement & Mechanism & Evidence \\
\midrule
1 & Console 3.5 mm analogue output is the only reliable booth interface; monitoring untouchable & Bridge console audio without degrading monitoring & USB audio bridge; analogue split, isolated monitoring branch & Dialogues \\
2 & Consumer audio software alters the signal in ways that corrupt a professional console feed & Clean pass-through; level control without a mixer & Consumer audio processing disabled at capture; non-linear gain control & Code audit \\
3 & Booth internet access is unreliable; events are confidentiality-bound & No in-booth cloud dependence & Local recognition plus language model modified to run on local hardware; automatic switch to local processing & Code audit \\
4 & Support arriving after my output is useless & Support within the interpreter's natural lag & Recognition exchanged between speakers without restart; a segment sent for translation only when recognizably complete; only recent context sent to the model & Code; telemetry \\
5 & Cutting audio at fixed time intervals breaks sentence structure; over-eager silence detection shatters words & Segmentation semantic-first, time-capped as fallback & Finalizes on sentence-ending punctuation; configurable pause threshold (live-adjustable only from 31 August); time cap; minimum-speech guard & Episode 1 \\
6 & Small models `parrot' under mixed-language noise; blunt script-filtering destroys acronyms & One-way direction, source-language terms unharmed & Letter-ratio language probe; prohibition instructions; near-duplicate screen & Episode 2 \\
7 & The display is read with peripheral vision during peak cognitive effort & Visual stability over expressiveness & Fixed left alignment; no letter-by-letter reveal; fixed layout; wheel zoom & Code audit \\
8 & Reference documents own the screen in document-heavy events & Overlay must coexist and never sleep & Two coexisting windows with system throttling neutralized; display-only companion & Episode 3 \\
9 & Term support must not compete for compute & Non-blocking terminology & Term matching in a separate worker; optional extraction in idle moments, off by default & Code audit \\
10 & The laptop's graphics memory can hold only one recognition model at a time & One recognition model resident at a time & The application starts, oversees, and retires recognition services on switch & Code audit \\
11 & Venue electromagnetic environment plus ungrounded power injects ground-loop noise; the grounded development desk masks it & Clean monitoring and capture under field power & Ground-loop isolator; alternative USB-DAC; grounded adapter or battery-only operation (grounding resolved it after the window) & Episode 4; recordings \\
\bottomrule
\end{tabular}
\end{table}
}

Row 11's hum was first heard on 12 August, its hardware remedy tested at
the focal event. The knowledge is not all of one kind: rows 1 and 4
through 9 carry judgments formed in interpreting practice, rows 2 and 10
general technical knowledge acquired during the build, rows 3 and 11
conference-industry and venue-engineering knowledge; the claim of
tacitness attaches to the first group, all three flowing through one
specification channel.

\subsection{Episode 1: segmentation, the syntax-preserving
compromise}\label{episode-1-segmentation-the-syntax-preserving-compromise}

The first implementation followed the generic default, fixed time
slices. In test replays of fast, connected speech it failed as any
interpreter could predict but no default anticipates. Slices severed
clauses mid-structure, and the model translated fragments without
syntactic context, once rendering a dependent clause as a full
assertion, a distortion more serious than omission. Silence detection
failed the other way, an aggressive threshold producing word fragments
that surfaced and vanished, and once relaxed, sentences stacking until a
timeout delivered `two or three sentences sutured together', as the log
puts it. Resolution came not from tuning but from re-specifying what
segmentation is for, in the requirement quoted in Section 3.3:
completeness, not clock time, releases a segment, and the time limit is
`a ceiling on a single pass, not a rhythm'. The founding mechanism
finalized on sentence-ending punctuation, with a fixed pause threshold,
a minimum-speech guard, and a hard cap whose sole function is to bound
worst-case latency. The pause control, live-adjustable since the v0.2
sprint, ships with per-tempo guidance from practice, roughly 0.4 seconds
for fast speakers to 1.5 for a deliberate one, expertise literally
embedded in the interface. Field use inside the founding window added a
second lesson. The focal event of Section 5 combined fast, interruptive
dialogue with long sentences, and genuine pauses proved scarcer than any
fixed threshold assumes, so that segments rode the hard cap, and the cap
is not merely syntactic, since segment length directly buys
translation-stage latency within the laptop's hardware limits for
graphics processing. Segmentation policy is thus simultaneously latency
policy, and the pause threshold became a live mid-event control in the
v0.2 sprint of 31 August, after the founding window had closed. The
knowledge at the bottom of the episode was never algorithmic, but the
professional judgment of what unit of speech a translation may receive,
and at whose tempo the boundary is set.

\subsection{Episode 2: hallucination governance, constraint from
intolerance}\label{episode-2-hallucination-governance-constraint-from-intolerance}

Under mixed-language input, source English against ambient Chinese
chatter, the local model exhibited what the logs call `parroting',
abandoning translation and copying the source. The executor's first
remedy was generic, filtering output by script, and the author rejected
it on professional grounds recorded in the dialogue almost in these
words: professional terms are often Latin-letter abbreviations, `and
they cannot be excluded' (translated), and a filter would erase
precisely the items the tool exists to support. The accepted design
encodes the requirement that direction is one-way and source-language
content survives only as terms. A lightweight letter-ratio probe,
weighing one Chinese character as three Latin letters, infers each
utterance's dominant language and overrides the direction setting on
disagreement, first articulated on the evening of 12 August. The model
receives prohibition instructions, never to reproduce the source
sentence, rather than after-the-fact filtering, and a screen-level check
discards any translation overlapping the source by at least 85 percent
after normalization, so that a failure of instruction becomes a failure
of display, not of the interpreter's trust. The unifying logic is that
intolerance of a failure mode, stated in operational terms, converts
into layered defense.

\subsection{Episodes 3 and 4: infrastructure and
hardware}\label{episodes-3-and-4-infrastructure-and-hardware}

Two briefer episodes recorded the same lesson beyond code. In
document-heavy events interpreters read the speaker's slides, and the
subtitle window froze whenever the operating system throttled the
process for losing focus, dying silently exactly when most needed; the
final architecture keeps two coexisting windows, neutralizes throttling,
and makes the subtitle a display-only companion incapable of touching
audio or models. No specification in the CAI literature lists `do not
suspend when unfocused'; it entered the list the day it failed in
rehearsal. And in the booth, from the first assignment on 12 August, the
monitoring headphones carried an audible hum no code could remove, while
the same chain was silent at home: the development desk draws grounded
power, the venue's charger did not, and the ground loop sang into the
analogue path. The remedy came in stages: an isolator and a second USB
audio adapter, bought on 17 August and tested at the focal event,
improved matters without removing the hum, and only the later
identification of grounding, met with a grounded adapter or battery-only
operation, silenced it. Both lessons share a register. The laboratory
hides what the working position reveals, and acceptance testing must
include the wall socket.

Across the episodes a stable division of labor emerges. The architect
supplied decomposition and tireless iteration, the executor turned
specification into behavior at a loop-sustaining pace, and what neither
supplied was the requirements themselves and the power of veto. In each
of the four narrated episodes the first proposal was the generic
default, and it failed against a constraint that lives only in practice.
A tally of the founding thread: of 101 architect exchanges, some thirty
concerned interface behavior, roughly double the fourteen on pipeline
logic, with twenty-five on the model stack, twenty on defects, nine on
terminology, five on genesis, categories overlapping slightly. The
knowledge load concentrated where ergonomics meets cognition, the
territory Section 2 predicted a display must defend. In Nonaka and
Takeuchi's (1995) terms, the dialogues are externalization events,
recording professional knowledge that had rarely needed to be told in
words before.

\section{5. In-booth use, an exploratory
case}\label{in-booth-use-an-exploratory-case}

This section examines the third research question on the evidence of a
single, demanding, real event, exploratory in a strict sense: no
baseline, no counterfactual self, no physiological measure, no efficacy
claims. Such a case can legitimately offer a precise account of how the
tool entered an expert's real-time strategy, structured by a human-led
principle: the interpretive work remains overwhelmingly the
interpreter's, the tool earning its place only in the narrow,
high-consequence slice where failure is unrecoverable.

\subsection{5.1 The setting}\label{the-setting}

The event was a three-hour installment of a publicly livestreamed
Chinese science-outreach interview series, broadcast on 18 August 2026,
in which a Chinese physicist-entrepreneur interviewed the Oxford
cosmologist Subir Sarkar on dark matter and dark energy, in rapid,
connected English with a marked accent. Three properties compounded. The
terminology was dense and partly outside the interpreter's active
knowledge. The accent shifted phonetic boundaries. And the material
contained minimal pairs whose confusion reverses meaning, most memorably
\emph{proton} and \emph{photon}. Six days earlier, on 12 August, the
author interpreted another installment with the tool's first live
deployment, a minimum viable version built that same morning, and rapid
speaker alternation exposed the manual direction switch as too slow; the
founding thread converted that failure into a requirement the same
evening. These events mark two early stages of the founding build: the
first re-specified the tool, the focal case ran the founding feature set
at its 18 August maturity.

\subsection{5.2 Three mechanisms of use}\label{three-mechanisms-of-use}

The three mechanisms below are the author's retrospective account of his
own strategies, interpretations of practice rather than measured
effects.

\textbf{Cross-modal verification.} Mid-interpretation, hearing alone
could not settle whether a speaker had said \emph{proton} or
\emph{photon}. Without support, the strategy is committed guesswork from
world knowledge, costly when wrong; with the tool, it changed modality.
On screen the recognized term appeared within a tolerable window, and a
glance settled what sustained listening could not. The perceived effect
was not that the interpreter knew more physics, but that a
micro-decision category migrated from high-effort guessing to low-effort
verification. The reflective notes describe this as the tool's most
valuable behavior, not translation, but certainty on demand where
certainty is scarce.

\textbf{Strategic lag management.} On syntactically dense sentences the
habitual economy is to stay close behind the speaker, minimizing memory
load at the price of processing pressure. With a trustworthy rendering
of the sentence tail stabilizing on screen, the interpreter deliberately
slowed and extended the lag on the hardest sentences, attending to the
front, where structure is decided, while the display held the tail. What
is attested is a retrospectively reported strategy: the reflective notes
record longer lags repeatedly chosen where they would not previously
have been risked.

\textbf{The safety net.} The least measurable effect was, subjectively,
the largest. An experienced interpreter can process almost anything, and
what erodes performance is the background allocation of attention to the
fear of missing something. Across the event, rotations included, that
background monitor stood down, on the felt assurance that a missed term
would reappear on screen within seconds. This is described as a
psychological safety net, not a measured one.

\begin{figure}[htbp]
\centering
\includegraphics[width=0.85\textwidth,height=0.85\textheight,keepaspectratio]{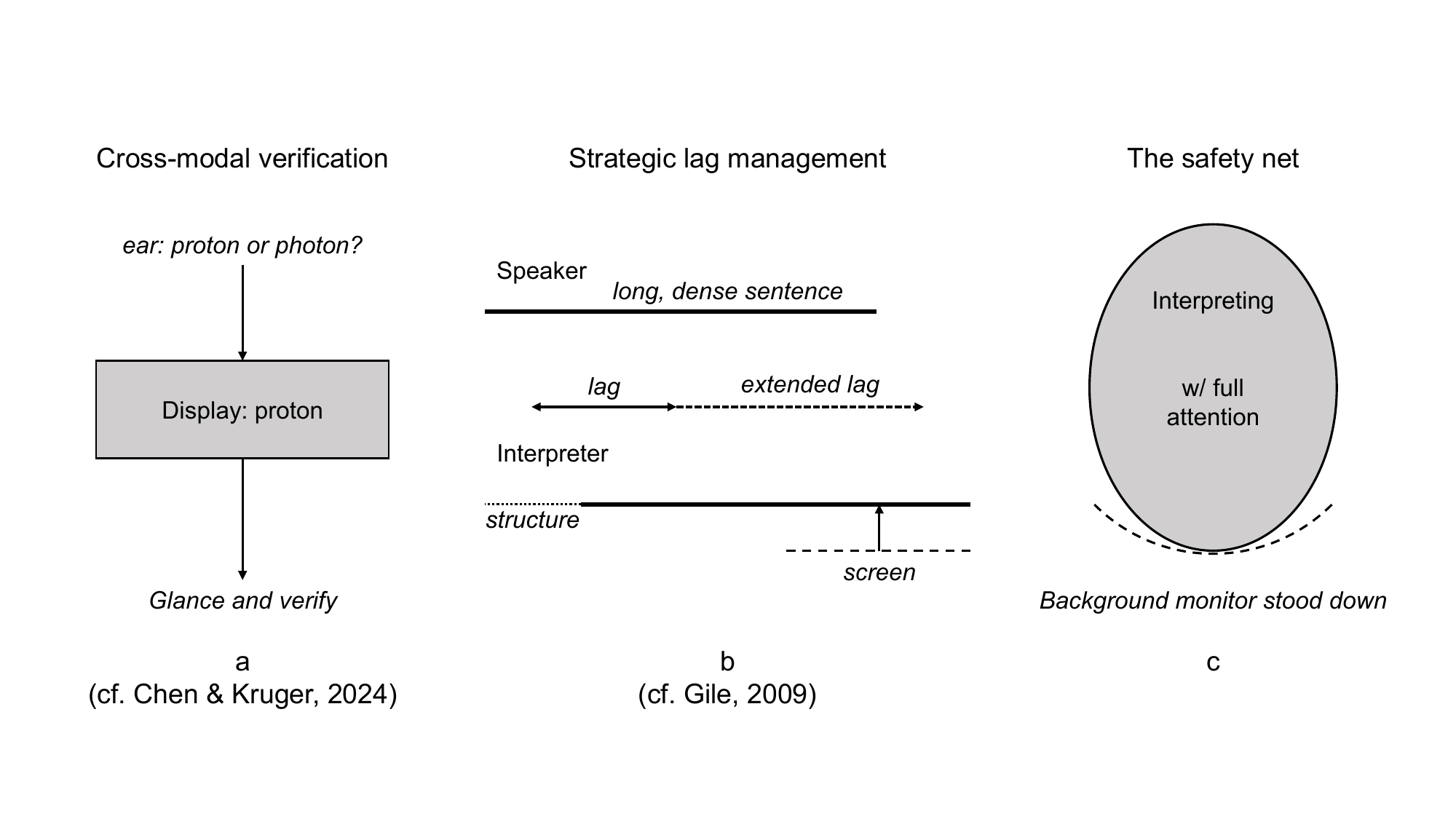}
\caption{Three mechanisms of in-booth use. Cross-modal
verification moves a micro-decision from guessing to verification when
the recognized term appears on screen. Strategic lag management extends
the lag on dense sentences, attending to structure at the front while
the display holds the tail. The safety net stands down the background
monitor for missed terms, freeing attention for interpreting.}
\end{figure}

The instrumented assignment of 28 August, a supplementary case distinct
from the focal event, preserves one such moment as a four-layer
reconstruction, two layers logged live and two recovered afterward, from
the event's one unscripted address, the English presentation by the
Egyptian embassy's commercial counselor.

what was said (session recording, manually re-transcribed for this
article): ``Egypt maintains the biggest FDI recipient in the region.''

what the display showed (recognition layer, live): ``Egypt maintains the
biggest F recipient in the region.''

what the machine translation produced (live):
``埃及仍然是该地区最大的外资接收国，但我们现在的目标是提高工业sector的占比''
(the abbreviation recovered in meaning; the machine layer translates
with recent context, and the second half renders surrounding content, in
mixed wording)

what the interpreter rendered (into Chinese, for the audience, recalled
by the author after the event):
``埃及一直是该地区最大的FDI接受国，即外国直接投资接受国'' (Egypt has
always been the region's largest FDI recipient, that is, of foreign
direct investment)

The display lost the abbreviation, the machine layer recovered its
meaning in Chinese while breaking its wording elsewhere, and the
interpreter's rendering kept the speaker's own abbreviation, carried by
neither on-screen layer, unpacking it for the audience. The moment
compresses the division of labor this article documents, an illustrative
reconstruction, not a synchronized record.

\subsection{5.3 Boundary and
corroboration}\label{boundary-and-corroboration}

What this case does not establish is that output quality improved, that
cognitive load decreased measurably, or that the tool transfers; the
focal event predates instrumentation, its account resting on reflective
notes, and corroboration exists in three forms. The first is the
instrumented assignment of 28 August (where some speaker scripts were
provided, as is standard practice), 971 recognized utterances, every one
translated, on a completely local system, at a median translation-stage
latency of 0.94 seconds (p90 2.16, p95 2.43, maximum 3.6 after excluding
a single 46-second cold start that remains in the denominator, Figure
5). Latency is the translation-stage log's own measure, from dispatch of
each translation request (segment text finalized) to completion of the
streamed response; capture, recognition, and rendering are excluded,
wait behind earlier requests is included, and the author's ear-voice
span was never measured, so arrival within the lag remains a design
goal, not a demonstrated property. All nine sessions ran the same local
chain, SenseVoice recognition and the same 7B translation model on one
laptop (August on the founding build, September on v0.2); the later
migration to a smaller model postdates this window.

The second is the focal event itself, whose public broadcast was
replayed on 1 September through a client replicating that day's chain at
default parameters, yielding 880 recognized utterances in which the
domain anchors appear, dark matter 83 times, neutrino 44, proton 6,
photon 5, the last two in genuine minimal-pair contexts. Human
verification then sampled the replay: six clips spanning the event's
timeline and speaking conditions plus every minimal-pair window, 118
utterances in which the author heard and judged 194 domain-term
occurrences against the broadcast audio, each scored captured or missed.
Capture stood at 57 percent of the judged occurrences, graded by
speaker, 60 percent under the guest's English, 45 percent in the host's
mixed Chinese--English; the proton-photon pair was captured in 11 of 13
occurrences, both failures producing the same substitution, `pluton'.
The damage sat not in core physics vocabulary, almost uniformly
captured, but in low-frequency proper names and multi-word terms. The
judgments are nested within clips and speakers and reported as a
description of this sample. Two boundaries follow: the capture
proportion indicates incomplete terminology coverage and is not directly
comparable with the subtitle-precision conditions examined by Li and
Chmiel (2024); the selective use of the display is described from
reflective accounts, not inferred from the rate. And the replay
validates recognition and term capture for this chain on this input, not
what the screen showed live on 18 August.

The third is instrumented use at scale: four further conference days (3,
4, 9, 11 September) added eight sessions and 5,274 translated
utterances; one day replicated the August profile with session medians
between 0.90 and 1.09 seconds, while a four-session marathon degraded to
a 3.38-second median before partial recovery on later days, so the
fresh-system figure is not a steady state under sustained load, and is
treated below.

\begin{figure}[htbp]
\centering
\includegraphics[width=0.85\textwidth,height=0.85\textheight,keepaspectratio]{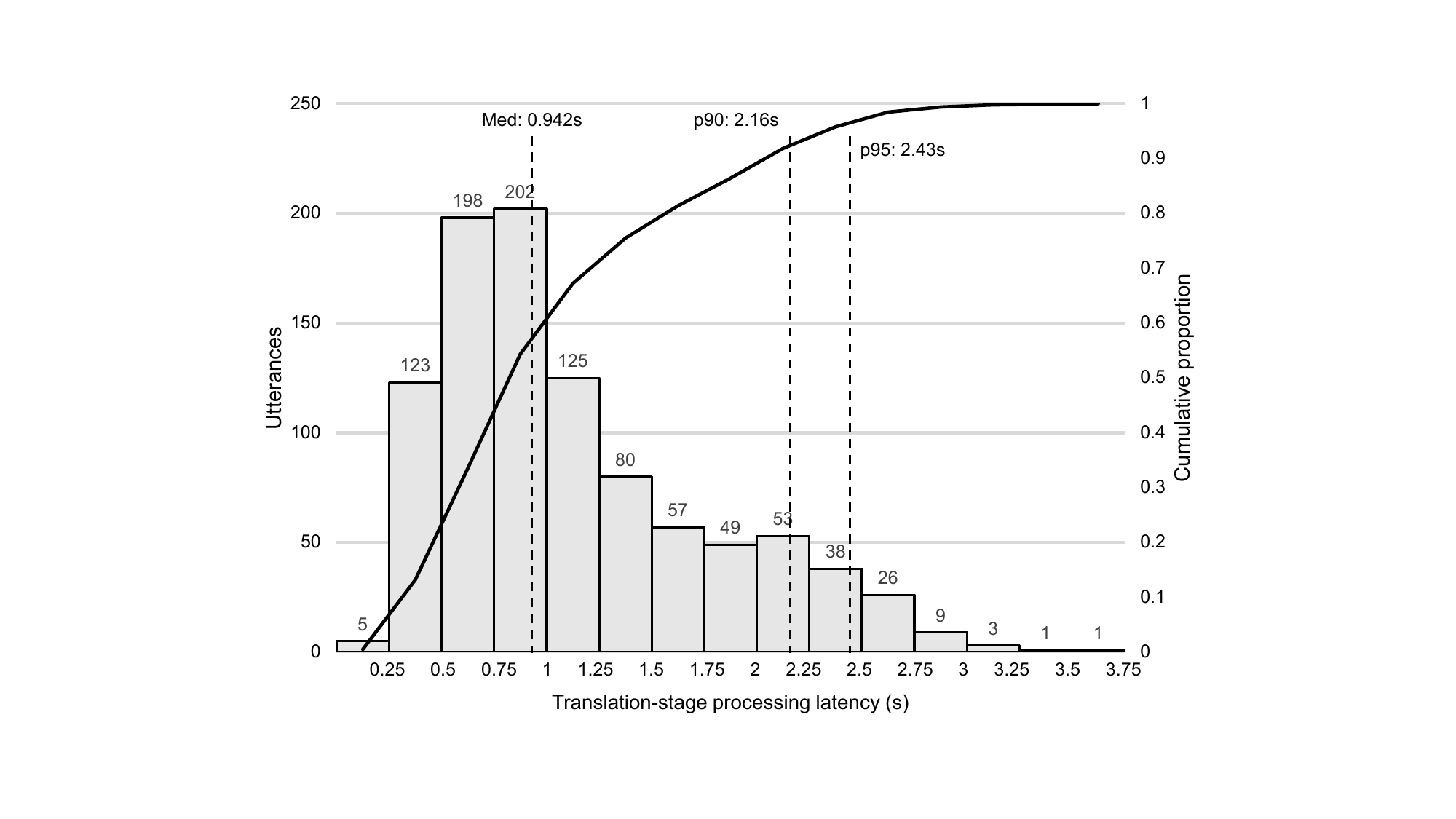}
\caption{Translation-stage processing latency (request dispatch
to streamed completion) on the instrumented assignment of 28 August (n =
971; the 46-second cold start remains in n). Histogram with cumulative
distribution; median 0.94 s, p90 2.16 s, p95 2.43 s.}
\end{figure}

The deviations inside the window did not end with it: pause scarcity
under fast dialogue and degradation under sustained load became the next
two months' development agenda, and segmentation evolved toward
punctuation as the primary basis, in routine use by late October. A
companion paper traces this evolution in detail (Zhao 2026); it
postdates and cites the present article, builds on the same telemetry,
whose 28 August figures originate here, and analyzes the post-founding
arc as a paradigm of practice-grown tools; the founding-period
externalization and in-booth mechanisms are analyzed only here.

Within these limits the case supports a restrained conclusion: in one
expert's hands, on one hard event, the tool served, deliberately and
repeatedly, as a targeted instrument of effort management, its uses
legible in advance from Section 2, since it was built to make them
possible.

\section{6. Discussion}\label{discussion}

This case speaks most directly to questions that translation studies has
made its own: where agency lies when language work is technologically
mediated, how professional knowledge is produced and carried, and where
the human-machine boundary should be drawn in accounts of the work. On
the first question the case offers a concrete reversal of the usual
direction of change, the practitioner moved, for once, from adopting
tools to defining them, and the remainder of this section reads that
reversal through the lenses that follow.

Read through lead-user theory (von Hippel 1986, 2005), the case is a
pure instance of user innovation: need-holder, specifier, acceptor, and
user were one person, and need-embedded information, costly to move,
never traveled. Three consequences generalize as conditional
propositions for practitioners building for their own constraints.
Requirements came from failure: each load-bearing feature of Table 2
originated in a concrete, experienced failure, so the strongest argument
for practitioner-led development is not that practitioners write better
software but that here the distance between experiencing a constraint
and encoding it was reduced to nearly nothing. The economics resolve
differently: Baldwin and von Hippel (2011) model when user innovation
displaces producer innovation, declining design and communication costs
chief among them, and conversational code generation is precisely such a
decline, a toolkit shifting design capability to users (von Hippel and
Katz 2002). And the artifact becomes a standing capability re-tunable
per assignment, against the one-size-fits-all logic of commercial tools.

The reversal also bears on how agency is described when humans and
machines share the work. O'Brien (2024) argues against antagonistic
dualisms that position the human and the machine as rivals for the same
tasks; the case adds a mechanism to that argument: here the division of
labor is not negotiated with a finished tool but authored into the tool
itself. Requirement origination, veto over generic defaults, and final
acceptance in the booth remained with the practitioner across every
workflow configuration; decomposition, iteration, and execution were
delegated. The four-layer reconstruction of Section 5.2 shows the same
division holding in operation, each layer failing differently, the
interpreter's rendering resting on neither machine layer alone. On this
evidence generative AI does not replace professional judgment; it gives
that judgment a new medium of expression, the specification, in which
judgment becomes architecture. Authorship in the strong sense, deciding
what the technology is for, stayed with the professional.

For CAI research, interpreter-built configurations are a new class of
object, likely to multiply as conversational code generation spreads, a
natural experiment in what interpreters believe technology should do for
them, and each such build is itself an act of knowledge production,
professional knowledge rendered into specification and archived in the
logs; the four-layer documentation of Section 5.2 offers a transferable
instrument for observing expert-AI collaboration. For the profession,
these abilities did not turn on programming as such: technical knowledge
was acquired just in time, and professional judgment guided that
learning. For interpreter and translator education, the case grounds
this in digital literacy: those same three abilities extend
interpreters' digital competence to building, not merely using, one's
own tools. These abilities are a starting inventory for curricula, a
teaching hypothesis rather than a derived framework, and the dialogue
logs are a ready-made case library.

The limitations are structural and stated plainly. It rests on one
researcher-practitioner, one language pair and market, and three
assignments over a single founding month, the September telemetry
serving as supplementary run data, not a second analysis window; its
cognitive claims are perceptual and unmeasured; the
developer-user-researcher identity cannot be anonymized away; and the
tool's computational load potentially limits battery life, though this
has not constrained the author's use. The obvious next study is
comparative, the same interpreter on comparable events with and without
the tool, alongside replication by others.

\section{7. Conclusion}\label{conclusion}

This article asked what happens when a conference interpreter, unserved
by the market and untrained in programming, becomes the developer of
their own tool. Three answers emerged. The appropriation proved feasible
through a triadic co-creation workflow, the expert's indispensable
contributions being requirement origination and acceptance judgment. The
process functioned as systematic externalization: knowledge about
segmentation, direction, latency, visual stability, and the physical
economics of the booth became written specification and then
architecture. And the tool entered real expert strategy in targeted,
legible ways, an effort-management instrument under a human-led division
of labor. The wider claim is deliberately modest: in this one
practitioner's experience, generative AI removed the practical barrier
between knowing what interpreting technology should do and being able to
make it do so. Whether it holds beyond this case is for the field to
test. The consequence for the profession is not replacement but
authorship (O'Brien 2024). For translation studies, the case suggests
that agency under technological mediation can be studied at the point of
construction, in the records of practitioners specifying what their
tools must do, and not only at the point of use. What interpreters build
next, for constraints only they know, is an empirical question for the
field, and this case offers one way to pursue it.

\subsection{Ethics statement}\label{ethics-statement}

The study analyzes the author's development dialogues and practice
records, the publicly broadcast focal event, and recordings and
telemetry from the author's contracted assignments, reported only in
aggregate or as annotated moments with no client-confidential content
reproduced. Speakers appear only in their public event roles; as an
autoethnographic study of the author's own records it required no
institutional ethics review.

\subsection{Funding}\label{funding}

This work was supported by an internal research grant of Capital Normal
University under its New Faculty Research Funding Program.

\subsection{Declaration of generative AI
use}\label{declaration-of-generative-ai-use}

The study's conception, research questions, professional knowledge,
analysis, and conclusions originate with the author. Generative AI was
used in two supporting capacities: as the object of study, an LLM
architect and coding agent implementing the tool under the author's
direction, and in manuscript preparation, GLM-5.3 (Z.ai) assisting with
language drafting, analysis scripting, and reference verification. The
author assessed, edited, and approved all AI-assisted content and takes
full responsibility for the content of the published article.

\subsection{Data availability}\label{data-availability}

The tool's source code is openly available at
https://github.com/omnigx/whisper\_interpreter (archived release,
https://doi.org/10.5281/zenodo.22255011). Development dialogue logs,
session telemetry, and replay analysis files are available from the
corresponding author on reasonable request. The offline guarantee
concerns client event audio and text in the sessions reported; it does
not extend to the development process itself, where AI services acted as
collaborators. The focal event's public broadcast is available at
https://tv.sohu.com/v/dXMvMTgzODM4MjcxMTkwLnNodG1s.html, and the 12
August installment at
http://tv.sohu.com/v/dXMvMTgyODUvNzM2ODQwNDExLnNodG1s.html.

\section{References}\label{references}

\begin{itemize}
\tightlist
\item
  Anderson, L. 2006. ``Analytic Autoethnography.'' \emph{Journal of
  Contemporary Ethnography} 35 (4): 373--395.
\item
  Baldwin, C., and E. von Hippel. 2011. ``Modeling a Paradigm Shift:
  From Producer Innovation to User and Open Collaborative Innovation.''
  \emph{Organization Science} 22 (6): 1399--1417.
  https://doi.org/10.1287/orsc.1100.0618.
\item
  Braun, V., and V. Clarke. 2006. ``Using Thematic Analysis in
  Psychology.'' \emph{Qualitative Research in Psychology} 3 (2):
  77--101.
\item
  Chan, V. 2025. ``Impact of Technology on Interpreting Practice: A
  Review of Studies on Technology and Interpreting Practice from 2013 to
  2024.'' \emph{Interactive Technology and Smart Education} 22 (1):
  81--102. https://doi.org/10.1108/ITSE-02-2024-0042.
\item
  Chang, H. 2008. \emph{Autoethnography as Method}. London: Routledge.
  https://doi.org/10.4324/9781315433370.
\item
  Chen, S., and J.-L. Kruger. 2023. ``The Effectiveness of
  Computer-Assisted Interpreting: A Preliminary Study Based on
  English--Chinese Consecutive Interpreting.'' \emph{Translation and
  Interpreting Studies} 18 (3): 399--420.
  https://doi.org/10.1075/tis.21036.che.
\item
  Chen, S., and J.-L. Kruger. 2024. ``Visual Processing during
  Computer-Assisted Consecutive Interpreting: Evidence from Eye
  Movements.'' \emph{Interpreting} 26 (2): 231--252.
  https://doi.org/10.1075/intp.00104.che.
\item
  Collard, C., and B. Defrancq. 2019. ``Predictors of Ear--Voice Span, a
  Corpus-Based Study with Special Reference to Sex.''
  \emph{Perspectives} 27 (3): 431--454.
  https://doi.org/10.1080/0907676X.2018.1553199.
\item
  Corpas Pastor, G. 2020. ``Language Technology for Interpreters: The
  VIP Project.'' \emph{Translating and the Computer} 42: 35--48.
\item
  Defrancq, B., and C. Fantinuoli. 2021. ``Automatic Speech Recognition
  in the Booth: Assessment of System Performance, Interpreters'
  Performances and Interactions in the Context of Numbers.''
  \emph{Target} 33 (1): 73--102.
  https://doi.org/10.1075/target.19166.def.
\item
  Fantinuoli, C. 2016. ``InterpretBank: Redefining Computer-Assisted
  Interpreting Tools.'' In \emph{Proceedings of Translating and the
  Computer 38}. London: AsLing. https://aclanthology.org/2016.tc-1.5/.
\item
  Fantinuoli, C. 2018. ``Interpreting and Technology: The Upcoming
  Technological Turn.'' In \emph{Interpreting and Technology}, edited by
  C. Fantinuoli, 1--12. Berlin: Language Science Press.
  https://doi.org/10.5281/zenodo.1493289.
\item
  Fantinuoli, C. 2023. ``Towards AI-Enhanced Computer-Assisted
  Interpreting.'' In \emph{Interpreting Technologies -- Current and
  Future Trends}, edited by G. Corpas Pastor and B. Defrancq, 46--71.
  Amsterdam: John Benjamins. https://doi.org/10.1075/ivitra.37.03fan.
\item
  Fantinuoli, C. 2026. ``Bridging the Usability Gap: Lessons from
  Interpreting Studies for Machine Interpreting Design.'' arXiv.
  https://arxiv.org/abs/2606.16009.
\item
  Fawzy, A., A. Tahir, and K. Blincoe. 2026. ``Vibe Coding in Practice:
  Motivations, Challenges, and a Future Outlook -- A Grey Literature
  Review.'' In \emph{Proceedings of the 2026 IEEE/ACM 48th International
  Conference on Software Engineering: Software Engineering in Practice
  Track (ICSE-SEIP)}, 212--223. New York: ACM.
  https://doi.org/10.1145/3786583.3786866.
\item
  Ge, Y., L. Mei, Z. Duan, T. Li, Y. Zheng, Y. Wang, L. Wang, et
  al.~2025. ``A Survey of Vibe Coding with Large Language Models.''
  arXiv. https://arxiv.org/abs/2510.12399.
\item
  Gile, D. 2009. \emph{Basic Concepts and Models for Interpreter and
  Translator Training}. 2nd ed.~Amsterdam: John Benjamins.
\item
  Hokkanen, S. 2017. ``Analyzing Personal Embodied Experiences:
  Autoethnography, Feelings, and Fieldwork.'' \emph{Translation \&
  Interpreting} 9 (1): 24--35.
  https://doi.org/10.12807/ti.109201.2017.a03.
\item
  Janikowski, P., and A. Chmiel. 2025. ``Ear--Voice Span in Simultaneous
  Interpreting: Text-Specific Factors, Interpreter-Specific Factors and
  Individual Variation.'' \emph{Interpreting} 27 (1): 28--51.
  https://doi.org/10.1075/intp.00116.jan.
\item
  Karpathy, A. 2025. ``There's a New Kind of Coding I Call `Vibe
  Coding'\ldots{}'' X (blog). February 2, 2025.
  https://x.com/karpathy/status/1886192184808149383.
\item
  Li, T., and A. Chmiel. 2024. ``Automatic Subtitles Increase Accuracy
  and Decrease Cognitive Load in Simultaneous Interpreting.''
  \emph{Interpreting} 26 (2): 253--281.
  https://doi.org/10.1075/intp.00111.li.
\item
  Mellinger, C. D. 2023. ``Embedding, Extending, and Distributing
  Interpreter Cognition with Technology.'' In \emph{Interpreting
  Technologies -- Current and Future Trends}, edited by G. Corpas Pastor
  and B. Defrancq, 195--216. Amsterdam: John Benjamins.
  https://doi.org/10.1075/ivitra.37.08mel.
\item
  Neustaedter, C., and P. Sengers. 2012. ``Autobiographical Design in
  HCI Research: Designing and Learning through Use-It-Yourself.'' In
  \emph{Proceedings of the Designing Interactive Systems Conference (DIS
  '12)}, 514--523. New York: ACM.
  https://doi.org/10.1145/2317956.2318034.
\item
  Nonaka, I., and H. Takeuchi. 1995. \emph{The Knowledge-Creating
  Company}. New York: Oxford University Press.
\item
  O'Brien, S. 2024. ``Human-Centered Augmented Translation: Against
  Antagonistic Dualisms.'' \emph{Perspectives} 32 (3): 391--406.
  https://doi.org/10.1080/0907676X.2023.2247423.
\item
  Pisani, E., and C. Fantinuoli. 2021. ``Measuring the Impact of
  Automatic Speech Recognition on Number Rendition in Simultaneous
  Interpreting.'' In \emph{Empirical Studies of Translation and
  Interpreting}, edited by C. Wang and B. Zheng, 181--197. London:
  Routledge. https://doi.org/10.4324/9781003017400-14.
\item
  Polanyi, M. 1966. \emph{The Tacit Dimension}. New York: Doubleday.
\item
  Prandi, B. 2023. \emph{Computer-Assisted Simultaneous Interpreting: A
  Cognitive-Experimental Study on Terminology}. Berlin: Language Science
  Press. https://doi.org/10.5281/zenodo.7143055.
\item
  Sarkar, A., and I. Drosos. 2025. ``Vibe Coding: Programming through
  Conversation with Artificial Intelligence.'' In \emph{Proceedings of
  the 36th Annual Conference of the Psychology of Programming Interest
  Group (PPIG 2025)}. https://arxiv.org/abs/2506.23253.
\item
  Seeber, K. G. 2011. ``Cognitive Load in Simultaneous Interpreting:
  Existing Theories --- New Models.'' \emph{Interpreting} 13 (2):
  176--204. https://doi.org/10.1075/intp.13.2.02see.
\item
  von Hippel, E. 1986. ``Lead Users: A Source of Novel Product
  Concepts.'' \emph{Management Science} 32 (7): 791--805.
\item
  von Hippel, E. 2005. \emph{Democratizing Innovation}. Cambridge, MA:
  MIT Press.
\item
  von Hippel, E., and R. Katz. 2002. ``Shifting Innovation to Users via
  Toolkits.'' \emph{Management Science} 48 (7): 821--833.
\item
  Zhao, Z. 2026. \emph{Growing One's AI Tools: A Paradigm of
  Practice-Grown Tools}. arXiv. https://arxiv.org/abs/2611.06132.
\end{itemize}

\end{CJK}
\end{document}